\documentclass[conference]{IEEEtran}
\usepackage{caption}
\IEEEoverridecommandlockouts
\usepackage{cite}
\usepackage{amsmath,amssymb,amsfonts}
\usepackage{algorithmic}
\usepackage{graphicx}
\usepackage{textcomp}
\usepackage{xcolor}
\usepackage{booktabs}
\usepackage{amsmath}
\usepackage{dsfont}
\usepackage{times}  % DO NOT CHANGE THIS
\usepackage{helvet}  % DO NOT CHANGE THIS
\usepackage{courier}  % DO NOT CHANGE THIS
\usepackage{newfloat}
\usepackage{listings}
\usepackage{amssymb}
\usepackage{xcolor}
\usepackage{float}
\usepackage{colortbl}
\def\BibTeX{{\rm B\kern-.05em{\sc i\kern-.025em b}\kern-.08em
    T\kern-.1667em\lower.7ex\hbox{E}\kern-.125emX}}
\begin{document}

\title{Bi-Grid Reconstruction for Image Anomaly Detection}

\author{
\IEEEauthorblockN{
Huichuan Huang\textsuperscript{},
Zhiqing Zhong\textsuperscript{},
Guangyu Wei\textsuperscript{},
Yonghao Wan\textsuperscript{},
Wenlong Sun\textsuperscript{},
Aimin Feng\textsuperscript{*}
}
\IEEEauthorblockA{
\textsuperscript {} Nanjing University of Aeronautics and Astronautics, China\\
MIIT Key Laboratory of Pattern Analysis and Machine Intelligence, China\\
\{hchuang, zhiqing, weiguangyu, wangyonghao, wenlong.sun, amfeng\}@nuaa.edu.cn
}
\thanks{*Corresponding author: Aimin Feng (amfeng@nuaa.edu.cn)} % 通讯作者说明
}

\maketitle

\begin{abstract}
% \normalfont
In image anomaly detection, significant advancements have been made using un- and self-supervised methods with datasets containing only normal samples. However, these approaches often struggle with fine-grained anomalies. This paper introduces \textbf{GRAD}: Bi-\textbf{G}rid \textbf{R}econstruction for Image \textbf{A}nomaly \textbf{D}etection, which employs two continuous grids to enhance anomaly detection from both normal and abnormal perspectives.  In this work: 1) Grids as feature repositories that improve generalization and mitigate the Identical Shortcut (IS) issue; 2) An abnormal feature grid that refines normal feature boundaries, boosting detection of fine-grained defects; 3) The Feature Block Paste (FBP) module, which synthesizes various anomalies at the feature level for quick abnormal grid deployment. GRAD's robust representation capabilities also allow it to handle multiple classes with a single model. Evaluations on datasets like MVTecAD, VisA, and GoodsAD show significant performance improvements in fine-grained anomaly detection. GRAD excels in overall accuracy and in discerning subtle differences, demonstrating its superiority over existing methods.
\end{abstract}


\section{Introduction}
\label{sec:intro}

Image anomaly detection and localization aim to identify and precisely segment abnormal regions in images, with applications spanning industrial inspection, medical imaging, and video surveillance. However, this task faces challenges due to the scarcity of abnormal samples and the diversity of anomaly patterns, ranging from minor scratches to significant structural damage in industrial production. Under these challenges, there is an increasing interest in developing unsupervised and self-supervised methods.

In anomaly detection, notable unsupervised methods include PaDiM \cite{padim}, SPADE \cite{SPADE}, and PatchCore \cite{PatchCore}, which utilize an external vector database to store features extracted from normal samples. During inference, anomalies are detected by calculating the Euclidean distance between the test sample and its nearest neighbor in the database. While effective, these methods face limitations due to their discrete feature storage, which hampers generalization and requires the retention of a large number of diverse normal features. This results in high spatial complexity and resource-intensive search operations.

\begin{figure}[t]
\centering
\includegraphics[width=0.48\textwidth]{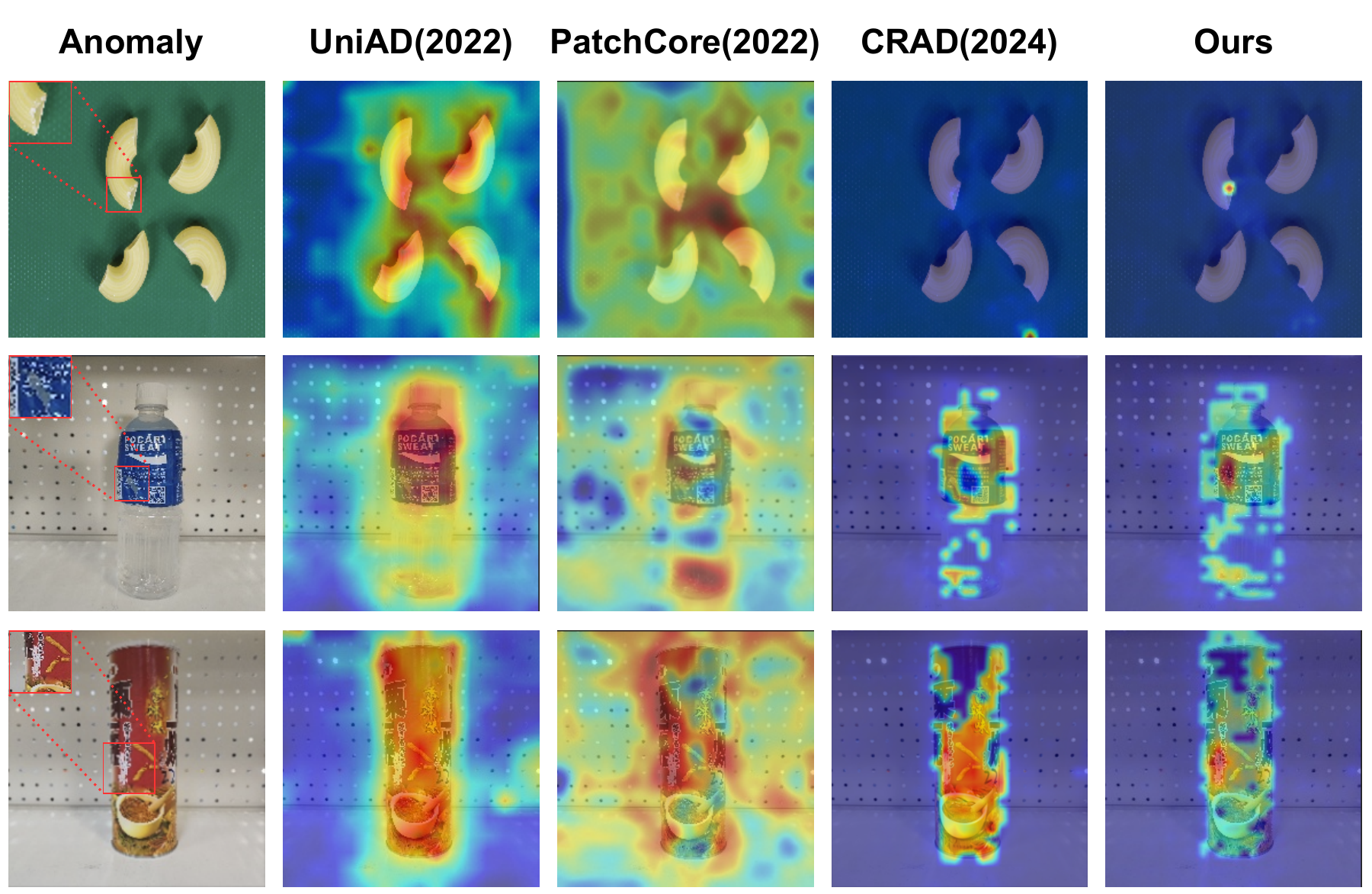} % Reduce the figure size so that it is slightly narrower than the column.
\caption{In the comparison of complex products and fine-grained anomalies, our model shows significant advantages over other models.}
\label{fig0}
\end{figure}

\begin{figure*}[ht]
\centering
\includegraphics[width=0.85\textwidth]{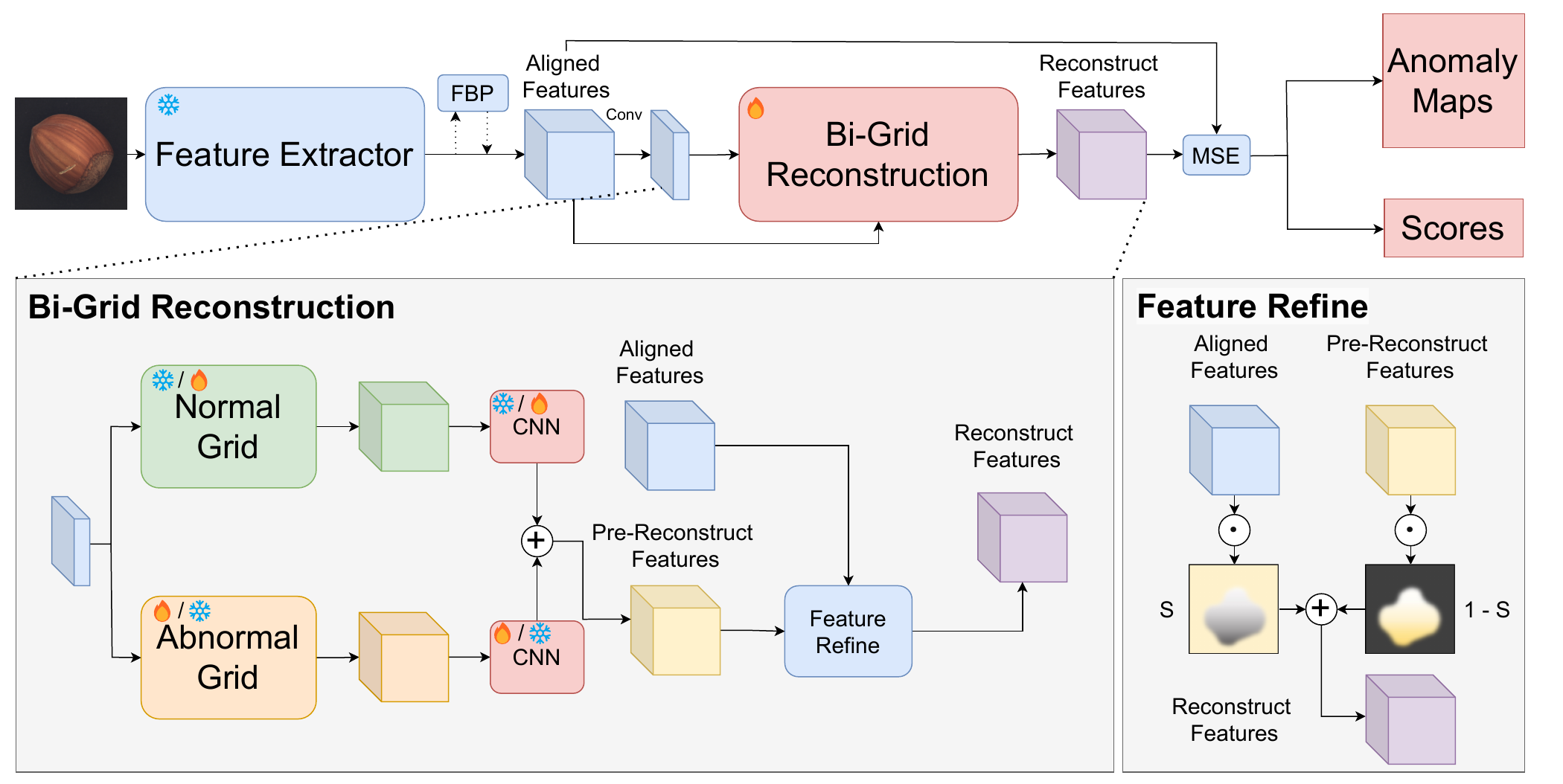}
\caption{\textbf{Overall framework of our GRAD.} The input samples are first processed by a pre-trained feature extractor to obtain initial features (the subsequent FBP module is only activated during the training of the anomaly grid). These features are then mapped to 2D coordinates through the coordinate mapping module. Based on these coordinates, sampling is performed from the normal and anomaly grid. The sampling results are fused and refined through the feature refinement module to produce the final reconstructed features. The comparison between these reconstructed features and the initial features yields the final anomaly detection results. (PS: The abnormal grid and normal grid have their top-left corner markers offset from each other, indicating that they also alternate during training.}
\label{fig2}
\end{figure*}

To address the shortcomings of insufficient generalization in the aforementioned methods, approaches such as MemAE \cite{MemAE} and DAAD \cite{DAAD} have been developed. These methods incorporate discrete repositories into the reconstruction task to generate generalized normal features and detect anomalies by comparing samples before and after reconstruction. By leveraging attention mechanisms, these models gather diverse normal features from the repository, resulting in stronger robustness to test data, and thus enhancing generalization performance. However, controlling the training of generative models remains a challenge. Overgeneralization can lead to the Identical Shortcut (IS) issue, where the input sample is mapped too closely to the reconstructed sample, as highlighted in UniAD \cite{UniAD}.

To balance generalization, CRAD \cite{CRAD} proposes using a continuous grid instead of discrete feature storage in the reconstruction task. Grid sampling improves generalization by using interpolation techniques, and compared to methods that rely on storing numerous features in memory, this approach reduces the risk of generating entirely new features (i.e., unseen anomalies in our context), thereby helping to avoid the IS issue. 

Although the aforementioned unsupervised methods have shown good performance, the boundaries of normal data they define often lack sufficient accuracy due to the absence of real anomaly data. This is particularly problematic when dealing with fine-grained defects, where over- or under-detection frequently occurs. To address the above issues, we propose GRAD, which introduces an anomaly grid that stores abnormal features in addition to the normal grid that stores normal features. This complements the knowledge learned from accessible synthetic anomalies, refining the boundaries of normal features, thereby enhancing the model's performance in detecting fine-grained anomalies in complex products. As shown in Figure \ref{fig0}, our model demonstrates significant improvements over previous methods in handling more complex products and fine-grained anomalies. Given that training models like DFMGAN \cite{DFMGAN} and AnomalyDiffusion \cite{anomalydiffusion} to synthesize realistic anomalies requires substantial computational resources, we also designed a Feature Block Pasting (FBP) module. This module synthesizes diverse anomalies at the feature level with controllable shape, size, intensity, and position to facilitate the rapid training of usable anomaly grid.

Our comprehensive analysis confirms GRAD as an effective AD solution, addressing limitations of existing methods and contributing to the integration of synthetic anomalies with unsupervised approaches. The main contributions of this paper are summarized as follows:
\begin{itemize}
  \item We propose a novel anomaly classification and localization method called GRAD. This method introduces an abnormal grid that incorporates knowledge from synthetic anomalies to refine the boundaries of normal features, significantly enhancing the detection performance for fine-grained anomalies.
  \item We design a lightweight method for anomaly synthesis at the feature level, called FBP, which allows flexible control over the location, size, intensity, and shape of synthetic anomalies.
  \item We tested GRAD on three image anomaly detection datasets: MVTec-AD\cite{MVTec}, VisA\cite{visa}, and GoodsAD\cite{GoodsAD}. The results show that GRAD achieves top-tier anomaly detection performance under a unified setting, overcoming the limitations of existing methods in detecting fine-grained anomalies.
\end{itemize}

\section{Related Work}
This section reviews various unsupervised anomaly detection methods, including reconstruction-based, embedding-based, and synthesis-based approaches.  Additionally, this section also emphasizes the effectiveness of grid feature sampling for reconstruction-based anomaly detection.

\subsection{Unsupervised Anomaly Detection}
Regarding the various unsupervised anomaly detection methods that have been proposed, they can be broadly categorized into three types:

\textbf{a) Reconstruction-based methods}: These methods assume models trained on normal samples reconstruct normal areas well but struggle with anomalies. Early efforts used various generative models like AE \cite{bergmann2018improving, chen2017outlier,zhou2017anomaly,dehaene2020iterative, liu2020towards}, GAN \cite{sabokrou2018adversarially, schlegl2019f, liang2023omni}, Transformer \cite{pirnay2021inpaintingtransformeranomalydetection, yao2023focusdiscrepancyintraintercorrelation}, and Diffusion Model \cite{diad} to learn the normal data distribution, attempt to replicate input data, and detect anomalies through reconstructing errors.

\textbf{b) Embedding-based methods}: These methods extract and store normal image representations from pre-trained networks, identifying anomalies via feature comparison. SPADE uses a multi-resolution semantic pyramid, PaDiM models the normal class with multivariate Gaussian distributions, and PatchCore employs greedy coreset subsampling for a memory-efficient approach. Anomalies are detected through feature cataloging and comparison.

\textbf{c) Synthesis-based methods}: These methods create anomalies on normal images, turning anomaly detection into supervised learning. CutPaste \cite{CutPaste} cuts and pastes patches randomly. DRÆM \cite{DRAEM} synthesizes pseudo anomalies using Perlin noise combined with out of distribution samples. NSA \cite{schluter2022natural} merges scaled patches with Poisson image editing. SimpleNet \cite{SimpleNet} adds Gaussian noise in the feature space. 

\subsection{Grid Feature Representation}
In the evolution of neural fields or neural representations, grid-based representations of signals parameterized by coordinate functions have proven effective across a range of applications, such as image and video processing \cite{gao2023fixedgridlearninggeometric}, 3D reconstruction \cite{jiang2020localimplicitgridrepresentations}, and novel view synthesis \cite{chen2022tensorftensorialradiancefields}. These grid structures efficiently capture high-frequency details without spectral bias and facilitate effective feature generalization through continuous feature spaces.

Considering the benefits of grid representation, CRAD uses continuous grid for anomaly detection, replacing discrete feature memory banks to improve generalization and address the Identical Shortcut problem. Furthermore, It combines global and local perspectives to capture structural features and detect anomalies across multiple classes, making it ideal for unified anomaly detection. In terms of computational complexity, the O(1) time complexity of grid calculations also surpasses the O(n) time complexity associated with discrete methods.

\section{Method}
GRAD mainly consists of a Feature Extractor, a Bi-Grid Reconstruction module, and an FBP module. In this section, we will explain these three components in sequence and provide details on our training and inference processes at the end.

\begin{figure}[t]
\centering
\includegraphics[width=0.45\textwidth]{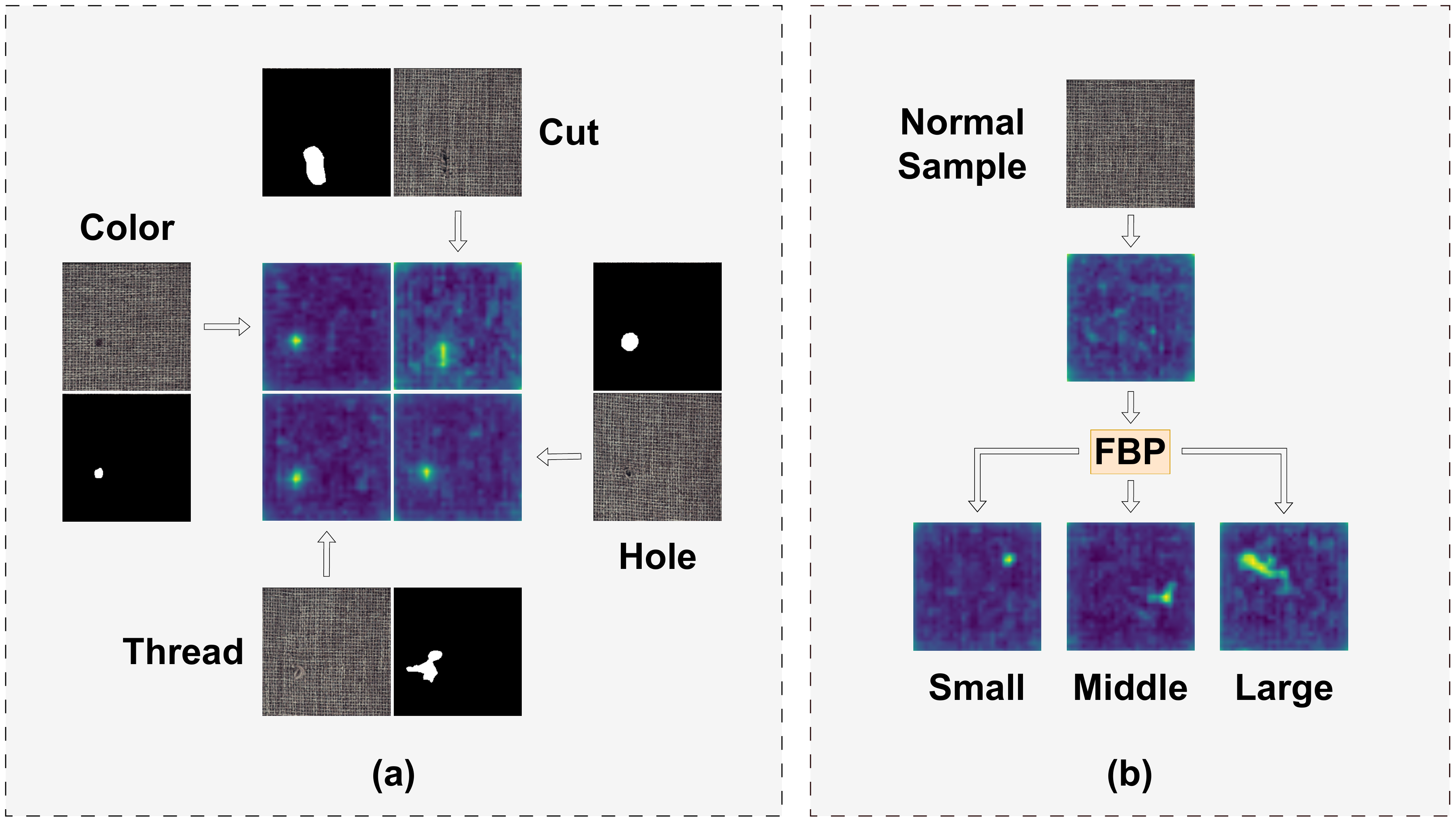} % Reduce the figure size so that it is slightly narrower than the column.
\caption{(a) Multiple anomaly patterns yield similar feature maps from the pretrained extractor. (b) Our FBP module can transform normal images into abnormal ones in the feature space.}
\label{fig3}
\vspace{-20pt}
\end{figure}

\begin{figure*}[t]
\centering
\includegraphics[width=0.82\textwidth]{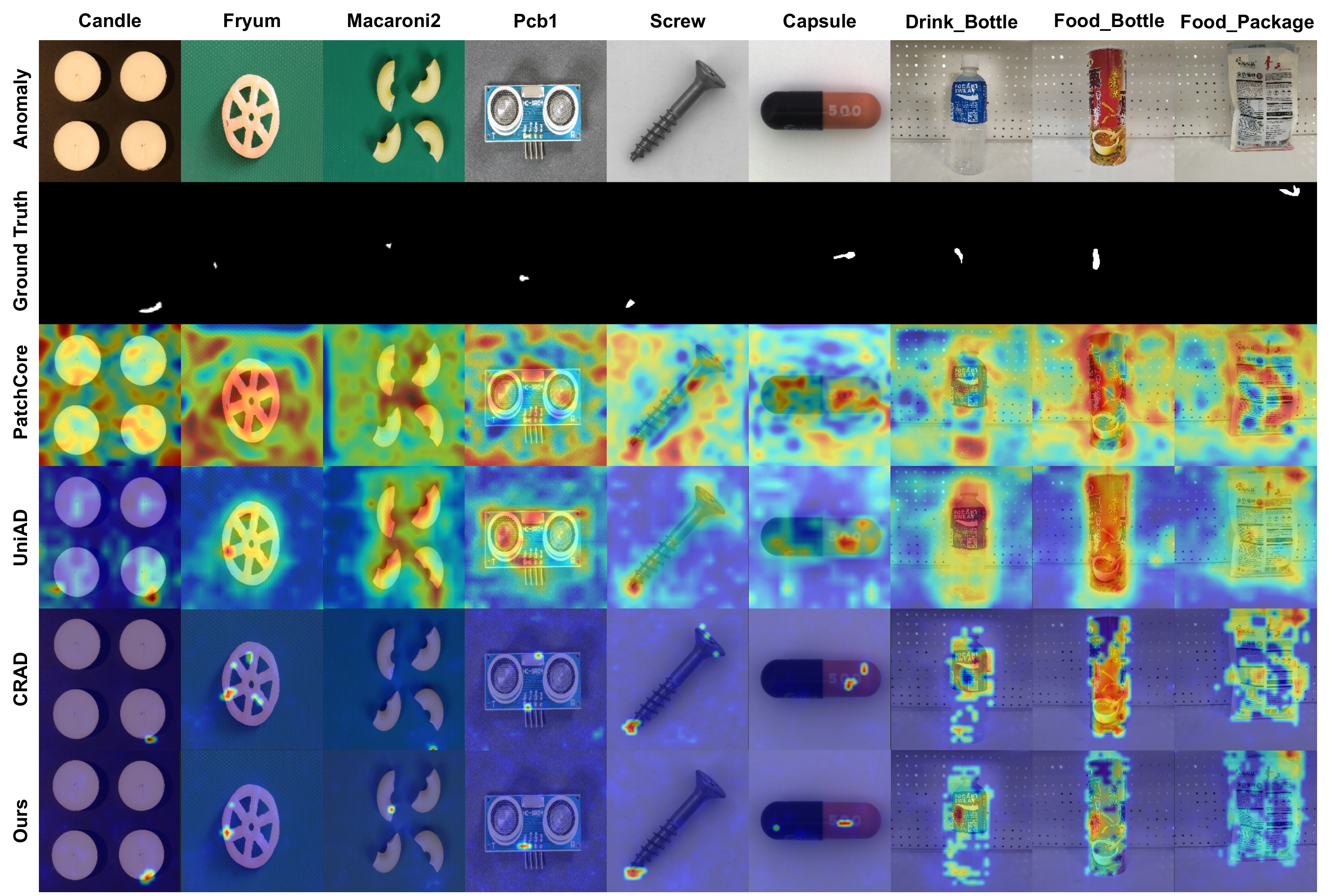} % Reduce the figure size so that it is slightly narrower than the column.
\caption{Qualitative results of GRAD on three datasets. Each row of the figure represents anomaly images, corresponding ground truths, results from different methods. Notably, even for extremely subtle anomalies in categories such as Macaroni2, Drink\_Bottle, and Food\_Bottle, our model has provided precise localization results.}
\label{fig4}
\end{figure*}

% GoodsAD Image-level
% unified
\begin{table*}[ht]
\centering
\begin{tabular}{l|c|c|c|c|c|c|c|c}
\multicolumn{9}{c}{\textbf{(a)} Image-level} \\
\toprule
Model & PaDiM & RIAD & DFR & UniAD & PatchCore & SimpleNet* & CRAD & Ours \\
\midrule
cigarette box & 81.0/80.8 & 67.2/70.1 & 52.1/58.5 & 94.5/93.8 & 92.9/93.7 & \textbf{96.2}/\textbf{96.9} & 93.0/93.7 & \underline{95.5}/\underline{96.2} \\
drink bottle & 53.9/55.3 & 62.3/60.9 & 47.8/52.9 & 65.7/66.6 & \underline{71.1}/\underline{75.1} & 61.8/74.0 & 70.6/72.5 & \textbf{74.5}/\textbf{77.8} \\
drink can & 53.3/55.3 & 62.1/59.3 & 54.9/54.0 & 60.0/63.9 & \underline{75.4}/\underline{78.2} & 74.7/60.6 & 73.5/74.7 & \textbf{81.6}/\textbf{84.6} \\
food bottle & 65.4/70.9 & 64.9/69.7 & 52.5/61.3 & 75.6/83.6 & 78.1/86.4 & 77.6/73.8 & \underline{83.0}/\underline{88.4} & \textbf{84.6}/\textbf{89.8} \\
food box & 58.5/71.7 & 57.6/71.0 & 53.2/67.1 & 62.3/73.1 & 67.6/73.2 & 63.3/66.4 & \underline{68.8}/\underline{73.8} & \textbf{70.2}/\textbf{75.7} \\
food package & 56.7/56.0 & 51.9/59.2 & 53.7/49.6 & 55.6/56.0 & 66.7/\textbf{64.9} & 59.7/62.9 & \underline{66.9}/62.2 & \textbf{70.1}/\underline{64.2} \\
\midrule
Mean & 61.5/64.8 & 61.0/65.0 & 52.4/57.2 & 69.0/72.8 & 75.3/\underline{78.6} & 72.2/71.5 & \underline{76.0}/77.6 & \textbf{79.4}/\textbf{81.4} \\
\midrule
\multicolumn{9}{c}{\textbf{(b)} Pixel-level} \\
\midrule
Model & PaDiM & RIAD & DFR & UniAD & PatchCore & SimpleNet* & CRAD & Ours \\
\midrule
cigarette box & 86.8/27.4 & 77.4/10.0 & 69.5/7.05 & 91.8/34.4 & \textbf{97.3}/54.8 & 96.1/\textbf{56.6} & 96.5/50.9 & \underline{97.0}/\underline{55.4} \\
drink bottle & 86.3/5.23 & 90.1/4.70 & 84.4/2.66 & 90.8/8.50 & 95.8/\textbf{39.2} & 92.0/10.7 & \underline{96.2}/24.9 & \textbf{97.5}/\underline{33.7} \\
drink can & 85.9/11.3 & 88.5/9.70 & 85.6/6.94 & 88.4/11.4 & 86.4/\underline{34.3} & 86.6/22.5 & \underline{95.2}/30.1 & \textbf{95.9}/\textbf{38.9} \\
food bottle & 89.3/10.1 & 90.4/8.10 & 85.1/4.30 & 93.7/20.1 & 94.8/48.6 & 94.8/37.8 & \underline{97.9}/\underline{52.9} & \textbf{98.3}/\textbf{58.6} \\
food box & 80.0/3.23 & 81.0/2.90 & 78.1/2.29 & 86.4/4.83 & \underline{92.8}/6.1 & 89.3/\underline{7.70} & 92.3/7.60 & \textbf{94.4}/\textbf{10.8} \\
food package & 84.9/2.78 & 89.4/2.90 & 85.9/2.18 & 89.6/3.63 & 94.6/\underline{17.2} & 90.2/7.30 & \underline{96.2}/13.5 & \textbf{97.6}/\textbf{21.0} \\
\midrule
Mean & 85.6/10.0 & 86.1/6.38 & 81.4/4.23 & 90.1/13.8 & 93.6/\underline{33.4} & 91.5/23.8 & \underline{95.7}/30.0 & \textbf{96.8}/\textbf{36.4} \\
\midrule
\end{tabular}
\caption{Image- and Pixel- level AUROC↑ / AUPR↑ on GoodsAD dataset, the * in the upper right corner of SimpleNet indicates that it is trained under the separated setting.}
\label{tab:data_comparison_all}
\end{table*}

\subsection{Feature Extractor} 
We define the feature extraction process as a preliminary step for subsequent work. Training and test sets are $\mathcal{X}_{Train}$ and $\mathcal{X}_{Test}$, respectively, with $\mathcal{X}_{Train}$ containing only normal samples and $\mathcal{X}_{Test}$ including both normal and abnormal samples. For a sample $x_i \in \mathbb{R}^{3 \times H \times W}$, we use a pre-trained EfficientNetb6 \cite{EfficientNet} on ImageNet to extract features $\Phi^i \sim \Phi(x_i)$. Due to data bias in the pre-trained network \cite{PatchCore}, we adapt it by selecting intermediate layers. For example, we select layers 3 and 4 from EfficientNetb6 layers 1 to 5, denoting them as $\phi^{l,i}$, where $l \in L=\{3,4\}$ represents the selected layers.

Next, we align the feature maps from different levels to the same size, and finally concatenate them along the channel dimension to obtain the aligned features for this stage:

\begin{equation}
    \phi_{aligned}(x_i)=f_{cat}(\{f_{resize}(\phi^{l,i},(H_{max},W_{max}))| l \in L \})
\end{equation}
 where $H_{max}$ and $W_{max}$ are the maximum height and width for all feature maps

\subsection{Bi-Grid Reconstruction}
The Bi-Grid Reconstruction includes a normal grid and an abnormal grid:

- \textbf{Normal Grid}: Trained solely on normal samples, this grid reconstructs input features into normal features via grid sampling. It addresses the Identical Shortcut (IS) problem by interpolating normal features for abnormal patches.

- \textbf{Abnormal Grid}: Trained with artificially synthesized anomalies or external anomaly samples, this grid helps refine normal feature boundaries during anomaly detection.

The normal grid captures both local and global features, while the abnormal grid categorizes features into normal and anomaly classes using masks and contrastive learning.We utilize the grid sampling method to obtain the following two features from the normal and abnormal grids, respectively: $\hat{x}_{i,n}$ and $\hat{x}_{i,a}$. Due to space limitations, please refer to \cite{CRAD} for specific details.

The features from two grids are fused through element-wise addition to produce preliminary reconstructed features:

\begin{equation}
    \hat{x}^{rec}_i=\lambda\hat{x}_{i,n} \oplus (1-\lambda)\hat{x}_{i,a}
\end{equation}

Here are some details regarding the training and inference of GRAD:

- \textbf{Training}: During the training phase, the normal grid of GRAD learns normal patterns through a reconstruction task using the Mean Squared Error (MSE) loss as the objective function:
\begin{equation}
    \mathcal{L}_{rec}=\frac{1}{CHW}||\phi_{aligned}(x_i)-\hat{x}^{rec}_i)||_2
\end{equation}
where $\phi_{aligned}(x)$ is the aligned feature of the input and $\hat{x}$ is the feature reconstructed by grid.

For the anomaly grid, we employ a contrastive learning idea to train it to increase the distance between the normal and anomaly features stored. We utilize the following truncated L1 loss:

\begin{equation}
    \mathcal{L}_{con} = \sum_{D^+} \frac{max(0, th-d^+)}{Len(D^+)} + \sum_{D^-} \frac{max(0, -th+d^-)}{Len(D^-)}
\end{equation}
where, $th$ is manually set to create a buffer zone around the separation boundary, with $th$ set to 0.5 in our experiments; $D^+$ and $D^-$ is a set of positive and negative sample pair  constructed via masks from FBP module or Ground\_Truth, where $d^+$ denotes the similarity of positive pairs and $d^-$ denotes that of negative pairs. The training of GRAD is conducted in two stages: initially, the anomaly grid is trained, followed by freezing the anomaly grid parameters and training the normal grid.

- \textbf{Inference}: The fused features from both grids are refined using a similarity-based feature refinement module to enhance detection confidence. The final anomaly score is obtained by comparing the reconstructed features with the original aligned features:

\begin{equation}
    pred=||\phi_{aligned}(x_i)-\hat{x}^{rec}_i||_2
\end{equation}

\subsection{Feature Block Paste}
The FBP module is designed to facilitate the rapid training of an anomaly grid that is ready for deployment. Compared to the method of adding Gaussian noise used by SimpleNet\cite{SimpleNet}, the anomalies synthesized using FBP are more diverse, resulting in a trained anomaly grid that achieves superior performance. We describe the FBP module as follows:

\begin{equation}
\phi_{pse\_ano}, mask = FBP(\phi_{nor}, M, B, I, P)
\end{equation}
where \( \phi_{pse\_ano} \) and \( mask \) represent the generated pseudo-anomalies and their corresponding annotation information, respectively. It takes five parameters: the feature map \( \phi_{nor} \) obtained from a normal image through the pretrained backbone, and the parameters M, B, I, P which control the shape, size, intensity, and position of the generated anomalies, respectively. 

Specifically, the FBP module operates by defining the block size \( B \), block intensity \( I \), and block center coordinates \( P = (x_c, y_c) \). We generate the initialization mask \( M \) as \( M = \text{zeros}(2B + 1, 2B + 1) \), a \( (2B + 1) \times (2B + 1) \) matrix initialized to zero. A random walk mask is created by selecting the initial position \( (x_0, y_0) = (B, B) \) and randomly choosing the number of steps \( N \) from \( [B, 2B] \). The random walk updates the position as \( (x_{k+1}, y_{k+1}) = (x_k + \Delta x, y_k + \Delta y) \) with \( \Delta x, \Delta y \in \{-1, 0, 1\} \), marking the corresponding position in \( M \) as 1. Finally, we initialize the block paste tensor \( P \) of size \( 1 \times 1 \times H \times W \) (initialized to zero), paste the block with intensity \( I \) at the marked positions in \( M \), apply Gaussian blur to obtain \( P_{\text{blurred}} \), and paste \( P_{\text{blurred}} \) onto the feature map.

\section{Experiments}

\subsection{Experiments Setup}
The methods used in our experiments follow a unified setup, where only one model is trained for all categories in the dataset, rather than a one-model-per-category approach, with the exception of SimpleNet.

\textbf{Datasets.} We assessed GRAD on three datasets: MVTec-AD, VisA, and GoodsAD. MVTec AD is a benchmark for industrial anomaly detection, VisA offers detailed pixel-level annotations for real-world scenarios, and GoodsAD focuses on anomalies in retail products, expanding the scope of anomaly detection to retail automation. Our experiments on these datasets evaluate methods' performance and adaptability across various contexts.

\noindent \textbf{Methods.} We assembled a benchmark of advanced unsupervised anomaly detection methods, spanning reconstruction-based, synthesizing-based, and embedding-based categories. The methods evaluated include PaDiM, RIAD \cite{RIAD}, DFR \cite{DFR}, UniAD, PatchCore, SimpleNet, and CRAD.

\noindent \textbf{Metrics.} Adhering to standard conventions, we employ both the Area Under the Receiver Operating Characteristics (AU-ROC/AUC) and the Area Under Precision-Recall (AUPR/AP) as metrics for assessing the performance of our models. 

\begin{figure}[ht]
\centering
\includegraphics[width=0.4\textwidth]{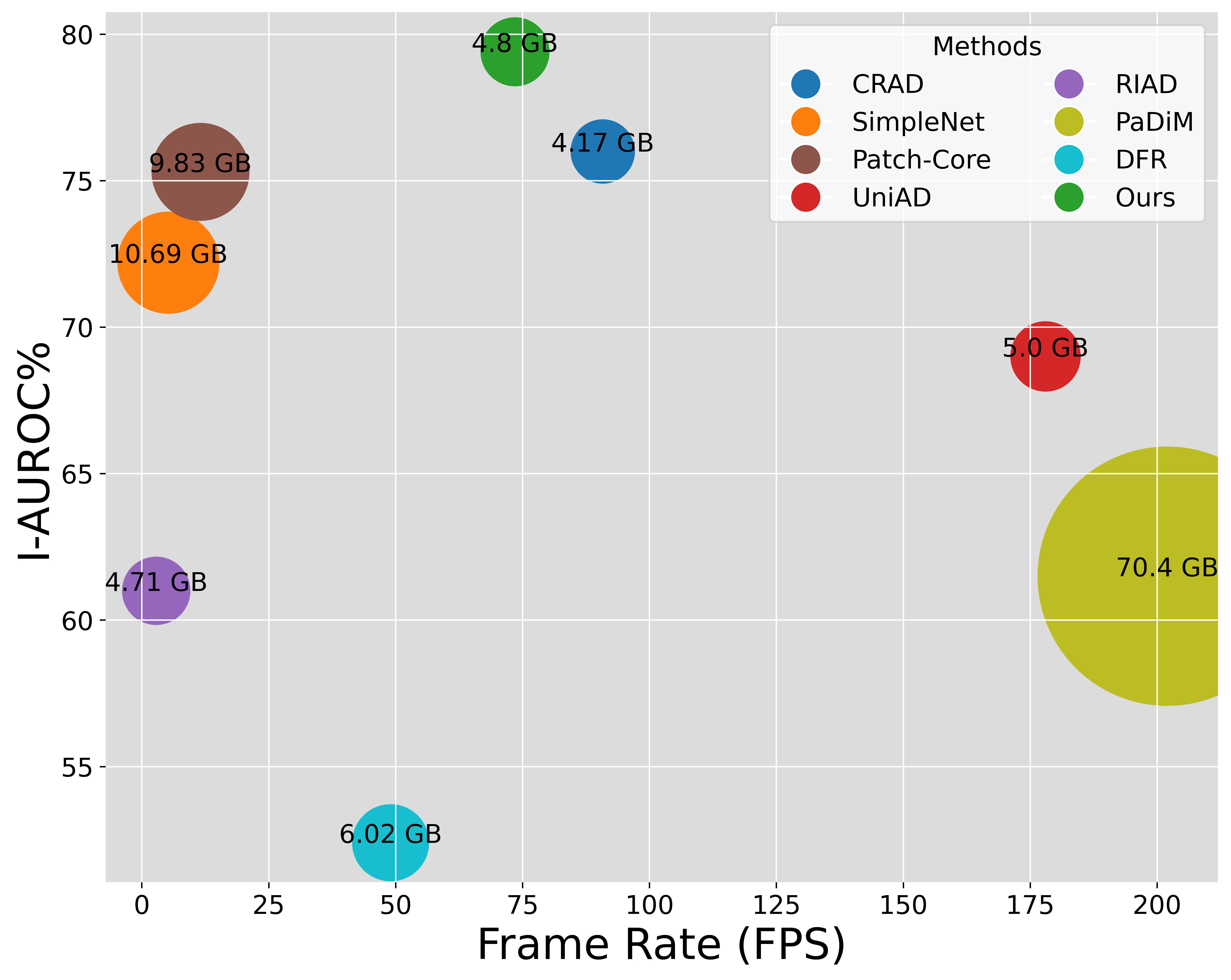} % Reduce the figure size so that it is slightly narrower than the column.
\caption{Comparing inference speed (FPS), I-AUROC, and memory occupancy on GoodsAD showcases the comprehensive performance of our model.}
\label{fig5}
\vspace{-20pt}
\end{figure}

\subsection{Comparison with Other Methods}
As shown in Table \ref{tab:data_comparison_all}, our model excels in both image-level and pixel-level evaluations on the GoodsAD dataset, achieving the highest mean AUROC 79.4\% \textcolor{red}{(+3.4\%)} and AUPR 81.4\% \textcolor{red}{(+2.8\%)} at the image level, and leading AUROC 96.8\% \textcolor{red}{(+1.1\%)} and AUPR 36.4\% \textcolor{red}{(+3.0\%)} at the pixel level. Additionally, the visualization results in Figure \ref{fig4} demonstrate that our model significantly outperforms existing methods in complex scenarios like GoodsAD and in detecting fine-grained defects. For more metrics and visualization results on MVTec-AD and VisA, please refer to the Appendix.

Additionally, we conducted a comprehensive comparison of GRAD and other methods in terms of memory usage and detection speed, as shown in Figure \ref{fig5}. Our model's memory usage is only 4.8 GB, significantly lower than PaDiM's 70.4 GB and SimpleNet's 10.69 GB. In terms of speed, our model achieves approximately 75 FPS, which is considered to be of moderate level. This indicates that our model not only outperforms other methods in detection accuracy but also maintains competitive spatio-temporal efficiency.

\subsection{Ablation Study}

\textbf{Normal-Grid and Abnormal-Grid.} We conducted an ablation study to assess the contribution of Normal-Grid (N-Grid) and Abnormal-Grid (A-Grid) to model performance. The results, summarized in Table \ref{tab:data_comparison_ablation}, indicate that the inclusion of both N-Grid and A-Grid yields the highest accuracy and recall on both the MVTec AD and Goods AD datasets. Specifically, the model achieved 99.8\% \textcolor{red}{(+0.8\%)} accuracy and 54.2\% \textcolor{red}{(+2.6\%)} Image/Pixel AUPR on MVTec AD, and 81.4\% \textcolor{red}{(+4.1\%)} and 36.4\% \textcolor{red}{(+6.6\%)} Image/Pixel AUPR on Goods AD when both grids were utilized. This indicates that the additional introduction of knowledge learned from synthetic anomalies beyond the normal grid can further enhance the performance of the model. For more ablation studies, please refer to the additional supplementary materials.

\begin{table}[H]
\centering
\begin{tabular}{cccc}
\toprule
N-Grid & A-Grid & MVTec-AD & GoodsAD \\ 
\midrule
$\checkmark$ & $\times$ & 99.0/51.6 & 77.3/29.8\\
$\checkmark$ & $\checkmark$ & \textbf{99.8}/\textbf{54.2} & \textbf{81.4}/\textbf{36.4}\\
\bottomrule
\end{tabular}
\caption{Ablation study for N-Grid and A-Grid}
\label{tab:data_comparison_ablation}
\end{table}

\section{Conclusion}
In this paper, we introduce GRAD, which incorporates an abnormal grid along with the FBP anomaly synthesis module, addressing the limitations of existing unsupervised and self-supervised methods in handling fine-grained anomalies. The success of GRAD lies in its use of an additional abnormal grid to refine the boundaries of normal features, and developing the Feature Block Paste (FBP) module for efficient and flexible anomaly synthesis at the feature level. Comprehensive experiments on industrial datasets such as MVTecAD, VisA, and the latest GoodsAD demonstrate GRAD's superior performance, improved detection accuracy.

\section*{Acknowledgment}
This work is supported by ‘the Fundamental Research Funds for Central Universities, NO.NJ2024031’.

\bibliographystyle{IEEEbib}
\bibliography{icme2025}

\end{document}